\documentclass{article}
\usepackage{iclr2027_conference,times}

\usepackage{amsmath,amsfonts,bm}

\def\eqref#1{equation~\ref{#1}}

\def\1{\bm{1}}

\DeclareMathAlphabet{\mathsfit}{\encodingdefault}{\sfdefault}{m}{sl}
\SetMathAlphabet{\mathsfit}{bold}{\encodingdefault}{\sfdefault}{bx}{n}

\usepackage{hyperref}
\usepackage{url}
\usepackage{graphicx}
\usepackage{booktabs}

\newcommand{\decoupage}{\textsc{D\'ecoupage}}

\title{\decoupage: Directorial Audio-Visual Control for Long Multi-Shot Video Generation}

\author{
\textbf{Yifan Wang}\textsuperscript{1,2},
\textbf{Hao Ouyang}\textsuperscript{2},
\textbf{Qiuyu Wang}\textsuperscript{2},
\textbf{Yihao Meng}\textsuperscript{1,2},
\textbf{Wen Wang}\textsuperscript{3},
\textbf{Zeyue Tian}\textsuperscript{1} \\
\textbf{Yichong Lu}\textsuperscript{1,2},
\textbf{Hanlin Wang}\textsuperscript{1,2},
\textbf{Qingyan Bai}\textsuperscript{1,2},
\textbf{Zichen Liu}\textsuperscript{1,2},
\textbf{Yujun Shen}\textsuperscript{2},
\textbf{Qifeng Chen}\textsuperscript{1}\\[0.6em]
{\small
\textsuperscript{1}The Hong Kong University of Science and Technology
\quad
\textsuperscript{2}Ant Group
\quad
\textsuperscript{3}Zhejiang University}
}

\iclrfinalcopy

\begin{document}

\maketitle
\vspace{-10px}
\begin{figure}[!ht]
    \centering
    \vspace{-12px}
    \includegraphics[width=\textwidth]{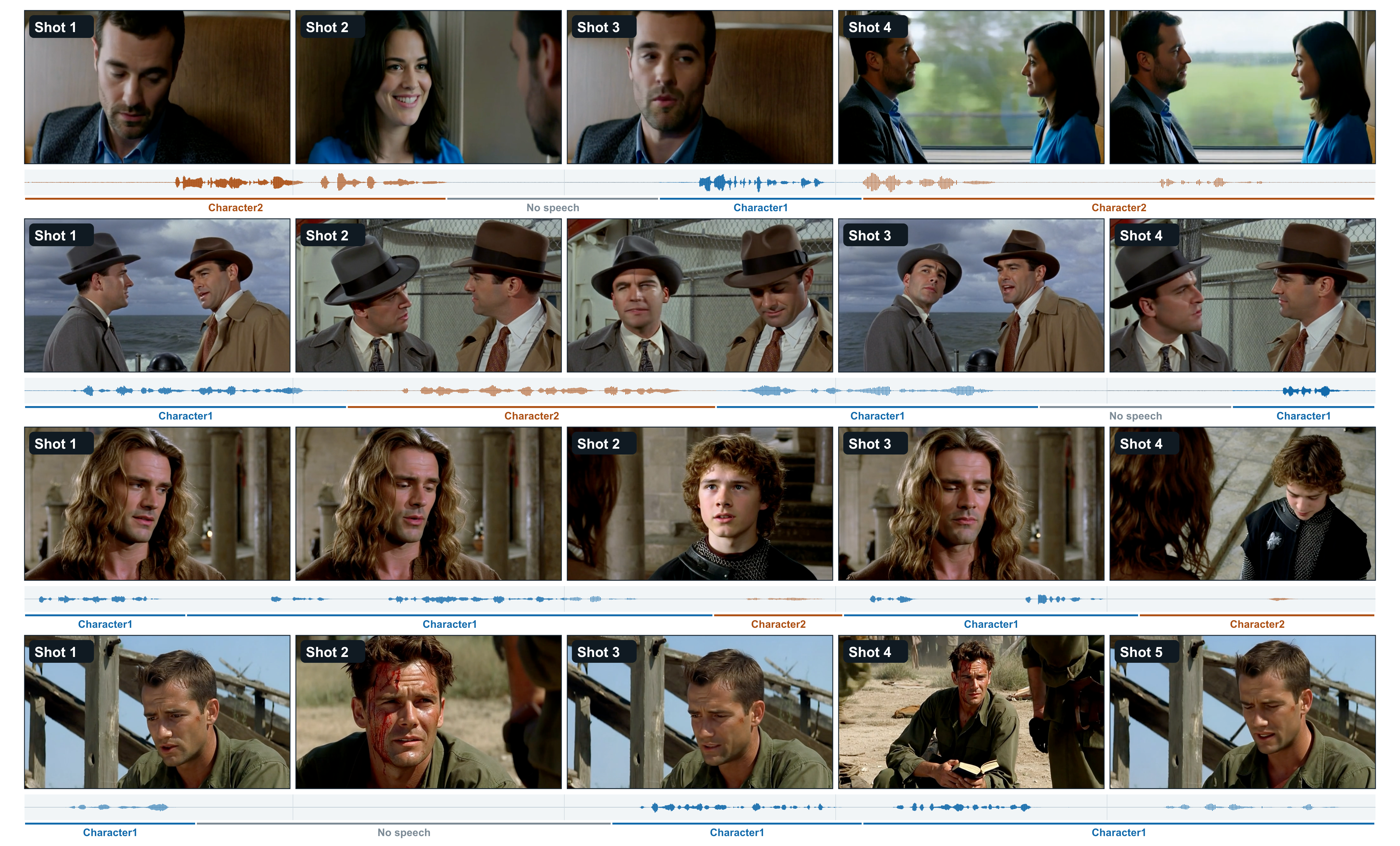}
    \caption{\textbf{Screenplay-directed audio-video generation.} 
    \decoupage{} directs visual and audio timing tracks across multi-shot scenes. Four 15-second examples demonstrate precise control over speech bridging across cuts (J-/L-cuts) and silent listener reactions, with all shot transitions and synthesized audio strictly following the screenplay timeline.}
    \label{fig:teaser}
    \vspace{-6px}
\end{figure}

\vspace{-5px}

\begin{abstract}
In cinematic storytelling, sound and visual transitions do not always coincide in lockstep: dialogue frequently bridges across cuts to reveal reactions or introduces new scenes via J- and L-cuts, while deliberate silences pace the scene. Existing audio-visual generative frameworks, however, lack explicit mechanisms to coordinate independent visual and acoustic timing over extended horizons. As narrative duration and shot complexity scale, these systems struggle to maintain long-range dialogue pacing—often hallucinating speech during intended silences or erroneously driving mouth movements on silent listeners. To enable fine-grained cinematic control, we present \decoupage{}, a screenplay-driven framework for long-form, multi-shot, and multi-character joint audio-video generation. Rather than coupling modality boundaries by default, \decoupage{} uncouples generation into independently bounded visual and acoustic tracks governed by persistent character identities. \decoupage{} builds on a joint audio-video diffusion model with dual-schedule temporal cross-attention, incorporating a directorial Lip-sync mechanism that selectively decouples speech conditioning from visual facial dynamics during off-screen dialogue. On multi-shot conversational benchmarks, \decoupage{} achieves precise control over shot transitions, audio timing, and visibility-aware lip synchronization, directing 15- to 60-second screenplay-driven audio-video generation with multimodal coherence. Demos are available on the \href{https://wyf0824.github.io/decoupage/}{project page}.
\end{abstract}

\section{Introduction}

\label{sec:introduction}

Cinematic storytelling relies fundamentally on the freedom to uncouple what is seen from what is heard. A camera cut marks a discrete visual shift, whereas dialogue routinely spills across visual boundaries: in J- and L-cuts, speech anticipates an incoming scene or carries past a cut to capture a listener's silent reaction. These expressive devices require independently scheduled visual and acoustic timelines, balanced by deliberate silences and tightly synchronized articulation whenever a visible character actually speaks.

Recent joint audio-video foundation models have enabled impressive synchronized multimodal synthesis~\citep{hacohen2026ltx2,liu2024syncflow,zhang2025uniavgen,ma2026ccl}. However, they inherently inherit a \emph{rigid synchrony assumption}—a built-in tendency to co-terminate shot boundaries, acoustic events, and facial dynamics within tightly coupled windows. In multi-shot conversations, this coupling falls short: dialogue easily drifts across cuts, scripted silences leak spurious speech, and off-screen voices erroneously trigger mouth movements on silent listeners. These failures expose two distinct directorial challenges: acoustic continuity and deliberate silences must be maintained independently across camera cuts, while active speech must animate only visible speakers. Crucially, extending visual continuity alone~\citep{meng2025holocine,meng2026causalcine} leaves these cross-modal pacing conflicts unresolved.

To address this gap, we introduce \decoupage{}, a screenplay-driven framework that brings fine-grained directorial control to multi-shot, multi-character audio-video generation. Rather than binding modality boundaries together, \decoupage{} decomposes generation into two independently timed visual and acoustic tracks aligned by a shared clock and persistent character identities. Under this formulation, dialogue can naturally span visual transitions while speakers move on- and off-screen. To support this decoupled design, we construct continuous 15- and 60-second training sequences by merging adjacent cinematic shots, pairing Whisper-derived speech timelines with Gemini-annotated character roles~\citep{radford2023whisper,klein2023fasterwhisper,reid2024gemini15} to natively retain sound bridges and reaction shots.

Building upon a joint diffusion backbone adapted via LoRA~\citep{hu2021lora}, \decoupage{} introduces two complementary mechanisms for decoupled control. First, dual-schedule temporal cross-attention enforces shot-aware video--text intervals and timestamp-aware audio--text intervals during denoising, steering multi-shot visual composition and acoustic dialogue schedules along independent timelines. Second, to resolve listener attribution errors, we introduce the directorial Lip-sync mechanism, an inference-time method that selectively regulates speech-driven articulation based on character visibility. During inference, whenever an utterance occurs off-screen, it selectively diverts the video stream's audio cross-attention to an auxiliary non-speech trajectory. This bypasses unwanted speech-driven facial articulation on reaction shots while retaining the main joint audio-generation path without requiring additional learned parameters.

Evaluating cinematic multi-shot conversations demands assessing both temporal fidelity and selective synchronization. Beyond standard generation metrics, a valid framework must verify that dialogue stays within its scripted window, that deliberate silences remain quiet, and that visible listeners stay quiescent during off-screen speech. Across 15- and 60-second conversational benchmarks, \decoupage{} substantially outperforms existing joint generators in shot-cut adherence, turn containment, and silence preservation, while effectively eliminating spurious lip synchronization on reaction shots.

Our contributions are threefold:
\begin{itemize}
    \item \textbf{Screenplay Formulation and Data Pipeline} We formulate multi-shot generation through independently bounded visual and acoustic tracks, supported by an automated pipeline that curates continuity-preserving 15- and 60-second scenes with persistent character identities.
    \item \textbf{Directorial Audio-Visual Generation}  We employ dual-schedule temporal cross-attention for independently timed shot and acoustic control, with a training-free directorial Lip-sync mechanism that reduces unintended listener articulation during off-screen dialogue.
    \item \textbf{Cinematic Benchmark and Scalability} We establish a screenplay-aligned benchmark with targeted metrics for cut fidelity, silence preservation, and listener quiescence. \decoupage{} achieves state-of-the-art script adherence on 15-second scenes and robustly sustains these directorial controls when scaled to minute-long narratives.
\end{itemize}
\section{Related Work}
\label{sec:related-work}

\subsection{Long-form and multi-shot video generation}

Long-form generation has increasingly separated narrative planning from visual
rendering. VideoDirectorGPT translates a prompt into an explicit scene plan
before generation~\citep{lin2024videodirectorgpt}, MovieDreamer combines a
hierarchical autoregressive narrative model with diffusion
rendering~\citep{zhao2025moviedreamer}, and StoryDiffusion introduces
cross-frame consistency mechanisms for image and video
stories~\citep{zhou2024storydiffusion}. VideoAuteur further uses rolling visual
context and keyframe-based rendering to produce long narrative
videos~\citep{xiao2025videoauteur}. These methods establish the value of
high-level story structure, but their plans do not directly specify an editable
soundtrack.

Another line makes the shot itself an explicit unit of control. ShotAdapter
models transitions with dedicated tokens and local attention
masks~\citep{kara2025shotadapter}; HoloCine combines shot-local text windows with
sparse inter-shot attention for minute-scale generation~\citep{meng2025holocine};
CausalCine retrieves relevant history during autoregressive shot
progression~\citep{meng2026causalcine}; and ShotDirector controls camera
parameters and editing patterns with shot-aware conditioning~\citep{wu2026shotdirector}.
These works primarily organize visual continuity; they do not directly model
the independently timed flow of dialogue across cuts. We instead treat the cut
schedule and the acoustic-event schedule as independently timed but jointly
generated tracks, so dialogue and sound can begin, end, or continue across a
visual edit.

\subsection{Joint audio-video generation}

Joint audio-video models replace post-hoc sound generation with coupled
denoising. SyncFlow uses interacting audio and video diffusion transformers for
text-conditioned synthesis~\citep{liu2024syncflow}; JavisDiT adds hierarchical
spatiotemporal priors for synchronized generation~\citep{liu2025javisdit}; and
Animate and Sound an Image jointly expands a still image into motion and
audio~\citep{wang2025animatesound}. More recent systems develop scalable
dual-stream architectures and stronger cross-modal interaction, including
LTX-2~\citep{hacohen2026ltx2}, UniAVGen~\citep{zhang2025uniavgen}, Cross-Modal
Context Learning~\citep{ma2026ccl}, and Harmony~\citep{hu2026harmony}.

Closed-source systems also support native joint generation. Seedance 1.5 Pro
uses a dual-branch diffusion transformer with cross-modal interaction~\citep{esser2024scaling}, while
Seedance 2.0 extends audio-video generation to multimodal reference and editing
inputs~\citep{seedance2025native,seedance2026world}.
Veo 3 jointly denoises temporal audio latents and spatiotemporal video
latents~\citep{deepmindveotech}.
Kling-Omni's technical report describes unified multimodal video generation
and editing~\citep{kling2025omni}; the subsequent Kling 3.0 release adds
native audio, multi-character dialogue, and shot-level storyboard
control~\citep{kuaishou2026kling3}. These capabilities make explicit
shot and dialogue control a relevant comparison axis, beyond perceptual
quality alone.

Controllable variants begin to expose speech and cinematic structure.
MM-Sonate adds phoneme-level control and reference-timbre conditioning for
intelligible, identity-preserving speech~\citep{qiang2026mmsonate}, while
CineDance and MAVIN extend joint generation toward multi-shot narratives with
structured annotations or hierarchical scripts~\citep{chen2026cinedance,liu2026mavin}.
These advances emphasize synchronized audiovisual coherence, but symmetric
cross-modal interaction alone does not specify when speech should animate a
visible face. In particular, an off-screen utterance and a listener's reaction
are temporally related without requiring lip synchronization. Our rigid
synchrony critique concerns this control gap, rather than an architectural
prohibition on off-screen sound. \decoupage{} makes shot and acoustic schedules independently addressable, and selectively regulates speech-driven articulation based on shot-level character visibility. We evaluate both temporal alignment and visibility-conditioned coupling explicitly.

\subsection{Multi-Character Speech and Cinematic Articulation}
\label{sec:related_multichar}

Audio-driven talking-face generation has evolved from single-speaker lip
synchronization~\citep{prajwal2020wav2lip,zhang2024musetalk,li2024latentsync}
to multi-character identity routing through spatial-temporal masks and
label-aware positional embeddings~\citep{kong2025multitalk,huang2025bind}.
These methods typically animate visible target faces from a supplied audio
signal. Their central task is to produce the intended articulation, whereas
cinematic dialogue also requires deciding when audible speech should leave
visible listeners unarticulated. Reaction shots and J- and L-cuts make this
distinction explicit: a speaker's voice can continue after their face has left
the frame. Conversational diagnostics likewise highlight the importance of speaker--listener attribution~\citep{zhou2026mtavg}. Our focus is this role decision within native joint generation. We leverage shot-level visibility to dynamically regulate the audio-to-video update, complementing asynchronous dialogue scheduling without requiring a separate talking-face synthesis stage.

\section{Method}
\label{sec:method}

We define a structured screenplay as two independently timed tracks specifying visual shots and acoustic events, with shared character identities and a common scene clock. Given this screenplay, \decoupage{} jointly generates a multi-shot video and its soundtrack. To follow these independent schedules,  we applies dual-schedule temporal cross-attention to the video and audio streams, allowing dialogue to span visual cuts. A training-free directorial Lip-sync mechanism further regulates audio-to-video cross-attention residuals based on shot-level speaker visibility, reducing unintended listener articulation during off-screen dialogue.

\subsection{Screenplay Formulation and Data Pipeline}
\label{sec:representation}

\paragraph{Dual-track representation.}
We represent the screenplay as a visual track
$\mathcal S_V=\{(I_i^V,c_i^V,\mathcal C_i)\}_{i=1}^{N_V}$
and an acoustic track
$\mathcal S_A=\{(I_j^A,c_j^A,\rho_j)\}_{j=1}^{N_A}$.
Here, $I_i^V$ is a shot interval, $c_i^V$ describes its framing
and action, and $\mathcal C_i$ contains its visible character
identities. Each acoustic event has an interval $I_j^A$ and
a description $c_j^A$ specifying dialogue or non-speech sound;
$\rho_j$ identifies the speaker when the event contains speech.
Character identities persist across shots and dialogue turns.

Both tracks use the same continuous scene clock $t\in[0,T]$, where $T$
is the sequence duration, but their boundaries are
independent. In particular, an acoustic event may overlap multiple
shots. Its speaker identity remains fixed, while its on-screen
or off-screen state changes according to the visible characters
in each overlapping shot. J- and L-cuts are thus specified by
placing acoustic events relative to visual cuts.

\begin{figure}[t]
    \vspace{-12pt}
    \centering
    \includegraphics[width=\linewidth]{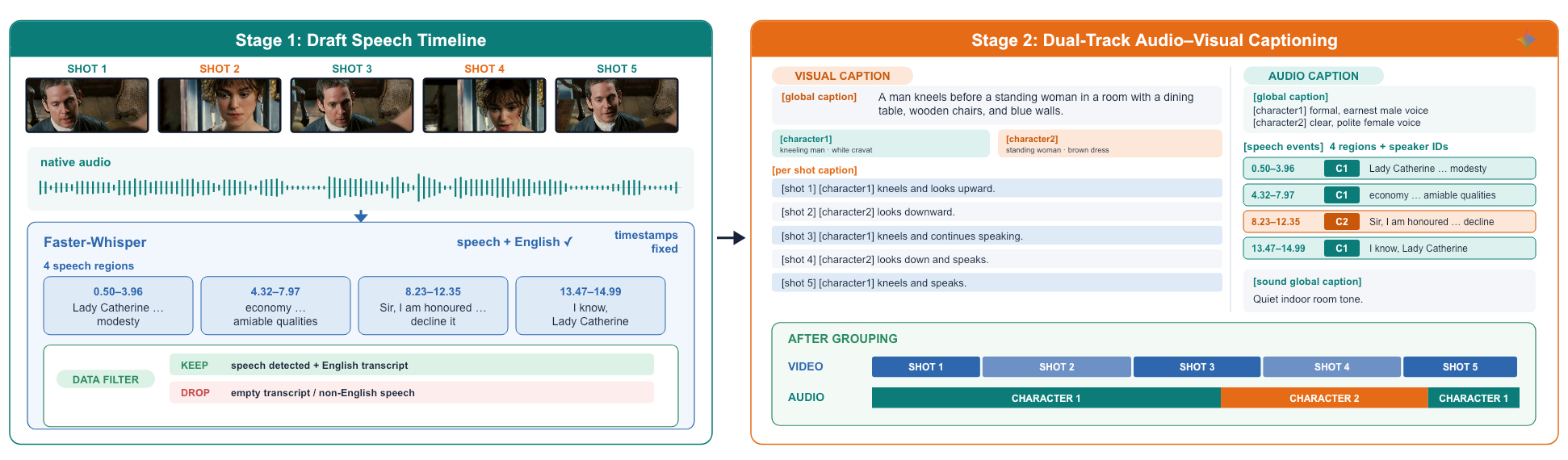}
    \caption{Joint audio-video data pipeline. Adjacent shots form audiovisual
sequences. Faster-Whisper establishes a fixed speech timeline and filters out
empty or non-English transcripts. Gemini 3.0 Pro then produces visual and audio
captions with persistent character identities. Finally, adjacent speech events
assigned to the same character are grouped into character-level audio intervals.
}
    \vspace{-10pt}
    \label{fig:data-pipeline}
\end{figure}

\paragraph{Continuity-preserving sequence construction.}
Following HoloCine's shot-based preprocessing~\citep{meng2025holocine},
we merge temporally adjacent shots from the same source into 15- and
60-second sequences. Video and native audio are resampled together,
preserving their alignment and dialogue across cuts. We remove duplicate
sequence plans and retain the shot boundaries.

\paragraph{Data captioning.}
We use Faster-Whisper~\citep{radford2023whisper,klein2023fasterwhisper}
to obtain transcripts and speech timestamps, which remain fixed during
annotation. Given cut-aware frame samples, the assembled audio, and
timestamped transcripts, Gemini 3.0 Pro produces visual and acoustic
tracks with shared character identities. The visual track describes
each shot's framing, visible characters, and actions. The acoustic track annotates dialogue and speaker identities along the Whisper timeline, marks intervals without speech as \texttt{<NO SPEECH>}, and uses a \texttt{[sound global caption]} to describe foley, ambient sounds, background music, and the scene's overall acoustic atmosphere.

Recurring characters receive persistent \texttt{[characterX]} identifiers
across shots and speech events, including off-screen dialogue; voices
whose speakers never appear in the sequence are labeled \texttt{[bg]}.
We group speech annotations by character and exclude audio annotations
involving more than four distinct speaking characters. We retain
reaction shots and off-screen dialogue rather than applying a global
lip-sync filter that could exclude these valid examples.

\begin{figure}[t]
\vspace{-10pt}
    \centering
    \includegraphics[width=\linewidth]
    {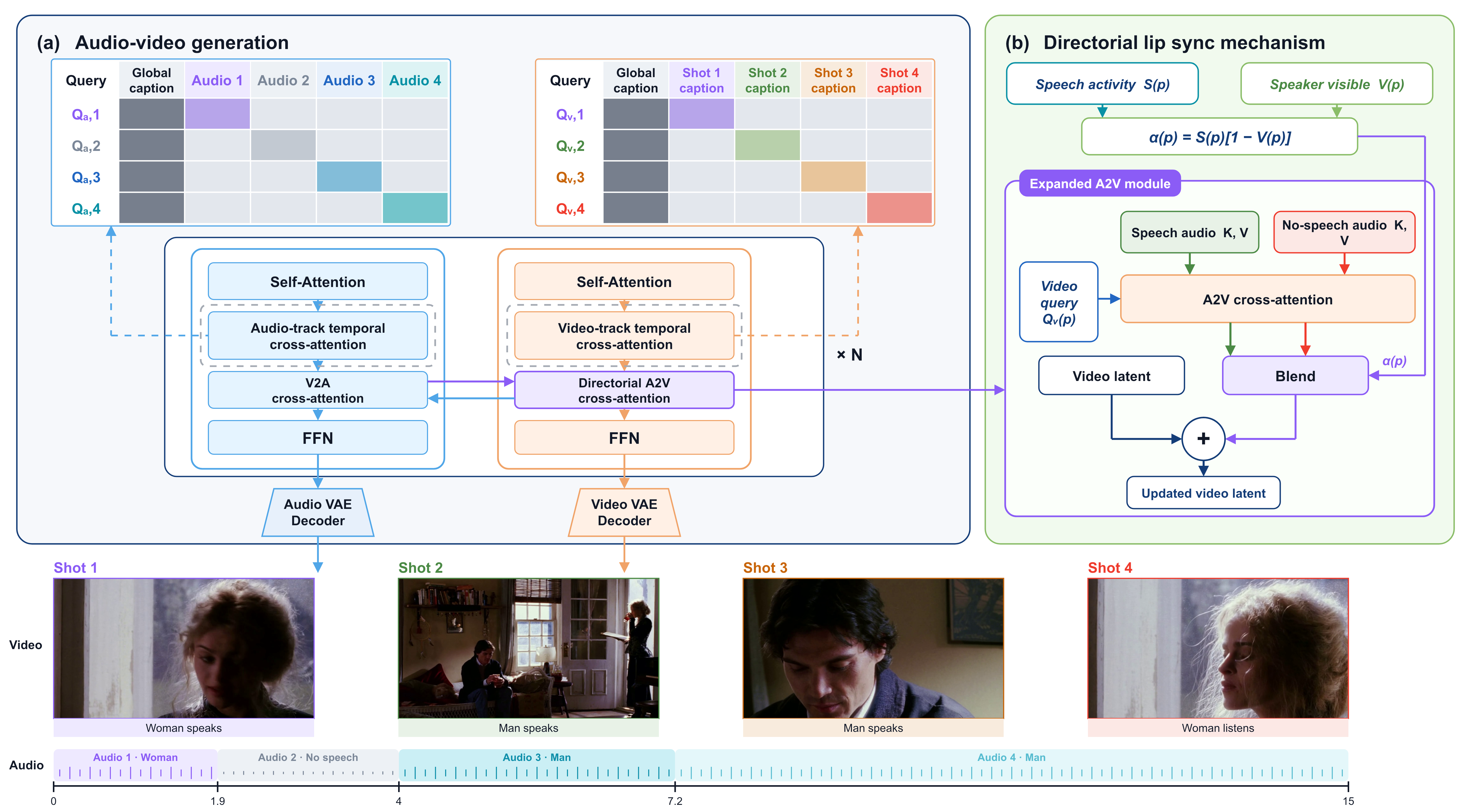}
    \caption{
Overview of \decoupage{}.
(a) Dual-track temporal cross-attention guides joint audio-video generation along independent visual and acoustic schedules, allowing dialogue to continue across shot boundaries.
(b) The directorial lip-sync mechanism blends speech and no-speech A2V residuals based on speaker visibility to suppress unintended listener articulation.
}
    \label{fig:method-pipeline}
\vspace{-8pt}
\end{figure}
\subsection{Dual-Schedule Temporal Cross-Attention}
\label{sec:routing}

To coordinate multi-shot visuals and continuous audio without enforcing rigid modality synchrony, we implement dual-schedule temporal cross-attention. Our dual-schedule isolates the visual and acoustic text conditions along their respective screenplay timelines: a visual cut shifts the accessible shot descriptions for video queries, while an ongoing speech or acoustic event conditioning remains intact across that transition. 

Specifically, for each modality $X \in \{V, A\}$, let $\tau_X(p)$ denote the scene timestamp corresponding to latent query token $p$. The textual conditioning keys are partitioned into global context tokens $\mathcal{G}$, shared register tokens $\mathcal{R}$, and interval-specific text tokens $\mathcal{T}_i^X$ associated with interval $I_i^X$. Notably, $\mathcal{G} = \mathcal{G}_V \cup \mathcal{G}_A$ simultaneously encapsulates sequence-level global descriptions for both video and audio. The accessible key index set $\mathcal{A}_X(p)$ for query $p$ is formally defined as:
\begin{equation}
\mathcal{A}_X(p)
=
\mathcal{G} \cup \mathcal{R} \cup \bigcup_{i:\,\tau_X(p) \in I_i^X} \mathcal{T}_i^X.
\label{eq:allowed-text-keys}
\end{equation}
For the visual track ($X=V$), each query token $p$ strictly attends to its current shot's prompt $I_i^V$. For the acoustic track ($X=A$), the union naturally accommodates multiple concurrent acoustic descriptions active at timestamp $\tau_A(p)$, seamlessly allowing ambient foley and environmental sounds.

We enforce this dual-schedule temporal cross-attention via an additive attention mask $M_{XT} \in \mathbb{R}^{N_X \times L_T}$:
\begin{equation}
M_{XT}[p, k] =
\begin{cases}
0, & k \in \mathcal{A}_X(p), \\
-\kappa, & \text{otherwise},
\end{cases}
\label{eq:text-routing-mask}
\end{equation}
where $\kappa$ is a large positive scalar ($e.g., 10^4$). The routed cross-attention is computed as:
\begin{equation}
\operatorname{Attn}_{XT} = \operatorname{softmax}\!\left(
\frac{Q_X K_{T,X}^{\top}}{\sqrt{d_X}} + M_{XT}
\right) V_{T,X},
\label{eq:routed-text-attention}
\end{equation}
where $Q_X$ denotes the latent queries of modality $X$, and $K_{T,X}, V_{T,X}$ are the text conditioning keys and values. Because recurring characters share consistent identifiers (\textit{e.g.}, \texttt{[character1]}) across different intervals $\mathcal{T}_i^A$ and $\mathcal{T}_i^V$, this dual-schedule formulation grounds long-range identity consistency across alternating visual shots and asynchronous dialogue turns.


\subsection{Directorial Lip-Sync Mechanism}
\label{sec:dlg}

While dual-schedule temporal cross-attention specifies \emph{when} camera cuts occur and \emph{when} dialogue begins or ends, it does not dictate \emph{whether} active speech should animate a visible character. During listener/reaction shots, dialogue naturally carries over while the camera captures a silent listener. Rather than leaving the video features unconstrained without acoustic guidance, we supply an explicit quiescent signal by deriving an auxiliary audio trajectory conditioned on \texttt{<NO SPEECH>}. This training-free mechanism selectively suppresses speech-driven lip motion during off-screen dialogue, while preserving the listener's natural facial expressions and emotional reactions undisturbed.
\paragraph{Paired speech and no-speech trajectories.}
At each denoising step, we obtain audio states from the native
joint trajectory and an auxiliary, paired-noise audio-only
trajectory conditioned on \texttt{<NO SPEECH>}.
At block $l$, these trajectories supply two sets of audio
keys and values,
$(K_{A,\mathrm{native}}^{(l)},V_{A,\mathrm{native}}^{(l)})$
and
$(K_{A,\mathrm{ns}}^{(l)},V_{A,\mathrm{ns}}^{(l)})$.
Both are evaluated against the same native video queries:
\begin{equation}
r_b^{(l)}
=
\mathcal F_{\mathrm{A2V}}^{(l)}
\left(
Q_V^{(l)},K_{A,b}^{(l)},V_{A,b}^{(l)}
\right),
\quad
b\in\{\mathrm{native},\mathrm{ns}\},
\label{eq:paired-a2v-residuals}
\end{equation}
where $\mathcal F_{\mathrm{A2V}}^{(l)}$ denotes the block's
audio-to-video cross attention and residual-output mapping.
The alternative residual therefore uses audio features from
the corresponding block and denoising step, rather than
a zeroed audio input or a separate full video prediction.

\paragraph{Directorial A2V residual blending.}
We blend the paired audio to video (A2V) residuals according to whether the
screenplay places the active speaker on screen.
For a video query $p$ in shot $i$, let $S(p)\in\{0,1\}$
indicate scheduled speech activity, and let
$V(p)=\mathbb{I}(\rho_j\in\mathcal{C}_i)$ indicate whether
the active speaker $\rho_j$ belongs to the set of characters
$\mathcal{C}_i$ shown in that shot.
These screenplay directives determine the blending weight
\begin{equation}
\alpha(p)=S(p)\bigl(1-V(p)\bigr),
\end{equation}
which we apply to the paired residuals at each block $l$:
\begin{equation}
\tilde r_{\mathrm{A2V}}^{(l)}(p)
=
\bigl(1-\alpha(p)\bigr)r_{\mathrm{native}}^{(l)}(p)
+
\alpha(p)r_{\mathrm{ns}}^{(l)}(p).
\label{eq:a2v-residual-mix}
\end{equation}
During off-screen dialogue ($S=1,V=0$), $\alpha=1$
replaces the native A2V residual with its no-speech
counterpart, providing quiescent acoustic guidance for
the visible listener.
When the speaker is on screen ($V=1$) or no speech is
scheduled ($S=0$), $\alpha=0$ retains the native residual.
The control is applied uniformly across video queries
within each temporal intersection of a shot and a speech
interval, requiring no spatial face masks.
At the A2V residual addition, the video state is updated as
\begin{equation}
h_{V,\mathrm{out}}^{(l)}(p)
=
h_{V,\mathrm{base}}^{(l)}(p)
+
\tilde r_{\mathrm{A2V}}^{(l)}(p).
\end{equation}
where $h_{V,\mathrm{base}}^{(l)}$ is the video state before
this residual is added. Subsequent block operations proceed
normally. This intervention changes the audio-derived update
received by the visible scene during off-screen speech,
while retaining native coupling for visible speakers.

The auxiliary audio trajectory requires no video backbone computation or video-to-audio (V2A) cross-attention, functioning solely to supply the no-speech audio states for the alternative A2V residual. 
Meanwhile, the main generation branch retains its native bidirectional coupling without directly replacing or blending the main branch’s audio prediction. 
This lightweight design introduces minimal computational overhead, as verified by the per-step runtime in Section~\ref{sec:inference-efficiency}.

\section{Experiments}
\label{sec:experiments}
\begin{table*}[t]
\vspace{-16pt}
  \centering
  \footnotesize
  \setlength{\tabcolsep}{0pt}
  \renewcommand{\arraystretch}{1.18}
  \begin{tabular*}{\textwidth}{@{\extracolsep{\fill}} l ccc cccccc c @{}}
    \toprule
    & \multicolumn{3}{c}{Video Quality} 
    & \multicolumn{4}{c}{Audio Quality} 
    & \multicolumn{2}{c}{Voice Identity} 
    & \multicolumn{1}{c}{AV Align.} \\
    \cmidrule(lr){2-4} \cmidrule(lr){5-8} \cmidrule(lr){9-10} \cmidrule(lr){11-11}
    Model 
    & Cut F1 $\uparrow$ & Cnt MAE $\downarrow$ & Imaging $\uparrow$
    & WER $\downarrow$ & Cont. $\uparrow$ & Leak $\downarrow$ & PQ $\uparrow$ & Consist. $\uparrow$ & Margin $\uparrow$
    & Sync-C $\uparrow$ \\
    \midrule
    \multicolumn{11}{@{}l}{\textbf{15-second Evaluation Dataset}} \\
    LTX-2.3 Base  & 0.26          & 1.23          & 0.68          & 35.69         & 52.68          & 18.69         & 6.18          & 0.42          & 0.12          & 2.52 \\
    MiniMax-H3    & 0.28          & 0.48          & \textbf{0.70} & 10.23         & 41.95          & 27.28         & \textbf{7.26} & 0.50          & 0.14          & \textbf{4.88} \\
    MOVA (I2VA)   & 0.18          & 7.38          & 0.56          & 38.22         & 39.33          & 28.47         & 6.26          & 0.36          & 0.01          & 3.24 \\
    \textbf{Ours} & \textbf{0.99} & \textbf{0.05} & 0.68          & \textbf{4.59} & \textbf{93.01} & \textbf{1.12} & 6.91          & \textbf{0.53} & \textbf{0.29} & 3.59 \\
    \midrule
    \multicolumn{11}{@{}l}{\textbf{60-second Evaluation Dataset}} \\
    LTX-2.3 Base  & 0.05          & 5.27          & 0.61          & 43.68         & 11.88          & 32.71         & 6.26          & 0.42          & 0.06          & 2.77 \\
    \textbf{Ours} & \textbf{0.76} & \textbf{1.81} & \textbf{0.66} & \textbf{7.13} & \textbf{91.12} & \textbf{4.46} & \textbf{6.84} & \textbf{0.57} & \textbf{0.22} & \textbf{4.00} \\
    \bottomrule
  \end{tabular*}
  \caption{\textbf{Quantitative results on the 15-second and 60-second benchmarks.} Metrics WER, Cont., and Leak are reported in percentage (\%). Bold font indicates the best performance within each duration group.}
  \label{tab:main-results}
\vspace{-10pt}
\end{table*}


\subsection{Datasets and Experimental Setup}
\label{sec:eval-setup}

\paragraph{Datasets and Evaluation Benchmark.}
The automated curation pipeline (Section~\ref{sec:representation}) yields approximately 30,000 cleaned, temporally aligned video--audio sequences for model adaptation. To benchmark performance under complex conversational dynamics, we construct an evaluation suite comprising 200 15-second and 200 60-second multi-character screenplays. Each screenplay explicitly specifies shot boundaries, recurring character identities, dialogue lines, and fine-grained acoustic timestamps, enabling standardized evaluation across both short- and long-horizon video generation.

\paragraph{Implementation Details and Baselines.}
We instantiate our approach upon the LTX-2.3 Base joint audio--video backbone~\citep{hacohen2026ltx2}, fine-tuning with LoRA (rank $r=32$, initial learning rate $5\times10^{-5}$ with a linear decay schedule) across 32 NVIDIA H20 GPUs. We compare \decoupage{} against representative open and proprietary joint audiovisual generation models, including the base model LTX2.3, MiniMax-H3~\citep{minimax2026h3}, and MOVA~\citep{siiopenmoss2026mova} in its image-to-video-audio (I2VA) configuration. All baselines are supplied with identical screenplay directives and evaluated using their official inference and sampling settings.

\subsection{Evaluation Metrics}
\label{sec:eval-metrics}

\paragraph{Cinematic Shot Control.}
We detect scene transitions at a fixed threshold of 0.18 and match them
one-to-one to screenplay cuts within 0.5 seconds. We report \textbf{Cut F1}
and the mean absolute error in cut count (\textbf{Count MAE}). In addition,
we evaluate the \textbf{Imaging Quality} dimension of VBench~\citep{huang2023vbench}
to assess overall video fidelity.

\paragraph{Dialogue and Pacing Control.}
We transcribe generated audio with Faster-Whisper~\citep{radford2023whisper,klein2023fasterwhisper}
and compute corpus-level \textbf{word error rate (WER)} against the reference
dialogue. \textbf{Containment} ($\uparrow$) is the percentage of target turns
whose aligned speech overflows the scripted window by no more than 0.2 seconds
for the 15-second evaluation dataset or 0.4 seconds for the 60-second evaluation
dataset. \textbf{Silence
Leakage} ($\downarrow$) is the duration-weighted overlap between recognized
words and the interior of scripted silence intervals. We assess acoustic
naturalness with \textbf{AudioBox P-Q}~\citep{vyas2023audiobox}. Exact
definitions and support rules appear in Appendix~\ref{app:evaluation-details}.

\paragraph{Vocal and Audiovisual Consistency.}
We evaluate character voice identity with ECAPA-TDNN embeddings~\citep{desplanques2020ecapa}.
\textbf{Role Consistency} is the cosine similarity of same-character voice embeddings across utterances; \textbf{Role Margin} is its within-video margin
over different-character similarity. \textbf{Sync-C} is the SyncNet confidence~\citep{chung2016syncnet}, which measures audiovisual alignment between a visible person's lip movements and the speech audio. To detect whether non-speaking characters are erroneously driven by off-screen dialogue, we introduce \textbf{TemporalBias}, which measures spurious audio-visual coupling on listener faces via time-shifted SyncNet contrast (details in Appendix~\ref{app:evaluation-details}). Lower values indicate that listener mouth movements remain natural and independent of acoustic timing, rather than being falsely modulated by speech.

\subsection{Quantitative Results}
\label{sec:quantitative-results}

\begin{figure*}[t]
    \vspace{-14pt}
    \centering
    \includegraphics[width=\textwidth]{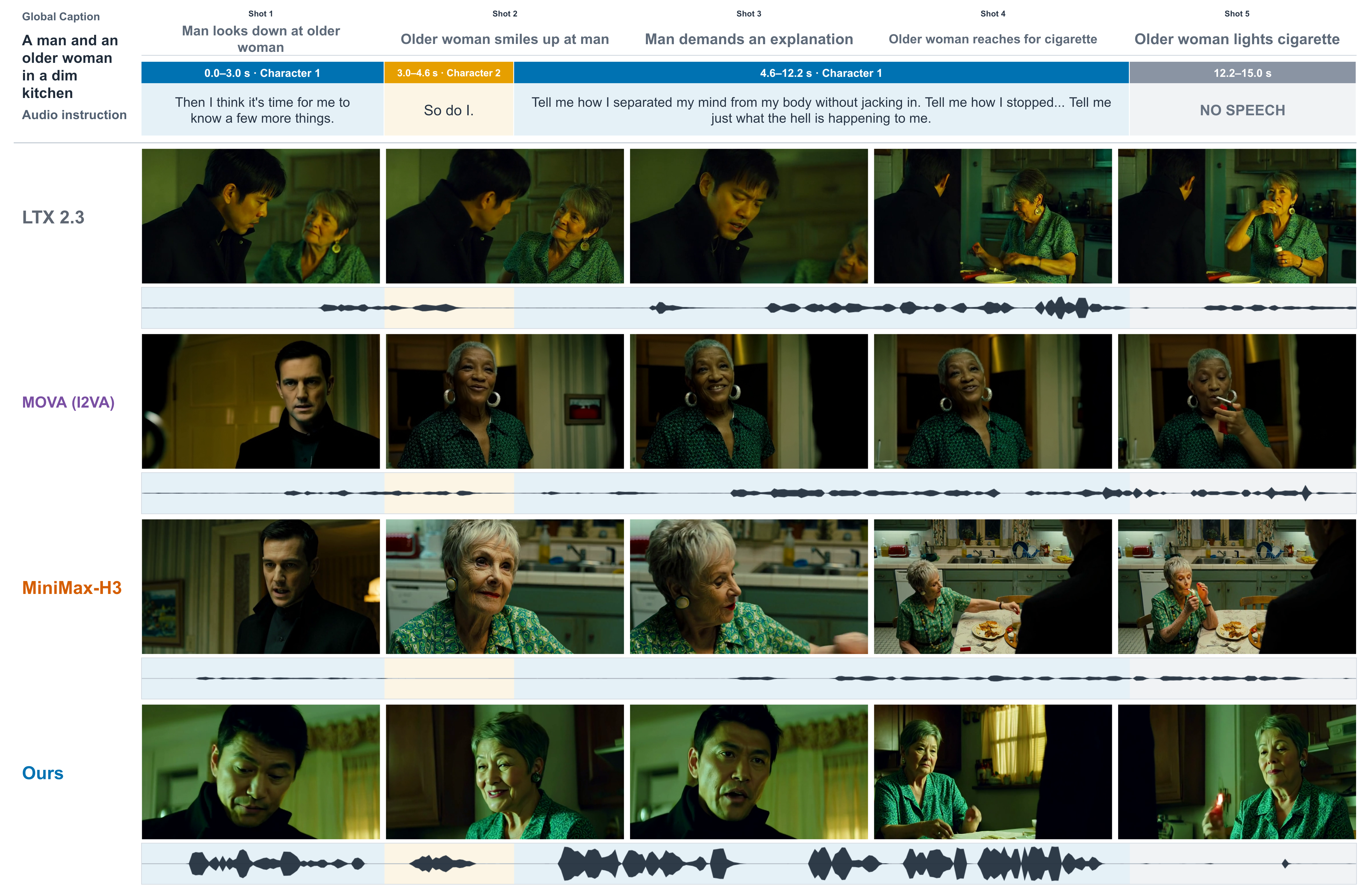}
    \caption{Qualitative comparison on a five-shot screenplay. Columns present shot prompts and the target dialogue schedule; rows show keyframes and synthesized audio waveforms across models. Blue/orange highlights mark each character's speech windows, and gray denotes silence. \decoupage{} adheres to the multi-shot structure across transitions, respects the Shot-4 listener, and correctly synthesizes the Shot-5 slience and lighter flick, achieving superior audio-visual alignment.}
    \vspace{-10.5pt}
    \label{fig:qualitative-comparison}
\end{figure*}

\paragraph{Comparison on 15-Second Generations.}
Table~\ref{tab:main-results} shows that \decoupage{} achieves
the strongest screenplay control and character voice consistency
among the evaluated methods on the 15-second benchmark.
It achieves near-perfect adherence to the requested edits, with
0.99 Cut F1 and 0.05 Count MAE.
It also obtains the lowest WER (4.59\%), the highest containment
(93.01\%), and the lowest silence leakage (1.12\%).
Beyond temporal control, \decoupage{} achieves the highest
Role Consistency (0.53) and Role Margin (0.29), maintaining consistent
character voices across conversations. These gains preserve competitive audiovisual quality. \decoupage{} also improves AudioBox P-Q from 6.18 to 6.91 and
Sync-C from 2.52 to 3.59. Although MiniMax-H3 achieves a higher Sync-C score (5.00),
the supplementary video comparisons show that \decoupage{} provides
satisfactory lip synchronization for the cinematic video generation.

\paragraph{Scaling to Minute-Long Narratives.}
When scaling the generation horizon to 60 seconds, \decoupage{} robustly maintains its high-fidelity temporal control and narrative coherence. As shown in Table~\ref{tab:main-results}, it preserves an impressive 0.76 Cut F1 and 91.12\% containment with a mere 4.46\% silence leakage. In stark contrast, standard baselines struggle severely over extended durations—for instance, LTX-2.3 Base degrades to 0.05 Cut F1 and 11.88\% containment, while its silence leakage surges to 32.71\%. Furthermore, \decoupage{} retains strong character voice consistency and separation, achieving 0.57 Role Consistency and 0.22 Role Margin. These results confirm that our framework's screenplay adherence and acoustic grounding scale gracefully to continuous, minute-long cinematic scenes without drifting or collapsing.

\subsection{Qualitative Results}
\label{sec:qualitative-results}

Figure~\ref{fig:qualitative-comparison} compares \decoupage{} with
LTX-2.3 Base~\citep{hacohen2026ltx2}, MOVA~\citep{siiopenmoss2026mova}, and
MiniMax-H3~\citep{minimax2026h3} on the same five-shot screenplay. The example
contains two alternating speakers, a long third dialogue turn that spans two
visual shots, and a final 2.8-second no-speech interval. The baseline rows show weaker adherence to the requested shot sequence
and poorer control of dialogue timing, consistent with the dataset-level
results in Table~\ref{tab:main-results}. In contrast, \decoupage{} faithfully follows the screenplay's visual-shot
and acoustic-event schedules: it reproduces all five requested compositions,
carries the third utterance across the cut to the Shot-4 listener without
driving the listener's lips, and preserves the Shot-5 no-speech interval
while generating the lighter flick. Qualitative results for
60-second generations are presented in Appendix~\ref{app:60s}, with
more qualitative examples provided in the supplementary materials.

\subsection{Ablation Study}
\label{sec:ablation-studies}
\label{sec:inference-efficiency}

Table~\ref{tab:ablation} examines both components on a fixed
subset of 136 of the 200 screenplays in the 15-second benchmark,
where the lip-sync mechanism is applicable. All ablation configurations
use the same subset, while Table~\ref{tab:main-results} reports
results on the full benchmark. Removing dual-schedule temporal
cross-attention reduces Cut F1 from 0.99 to 0.34 and dialogue
containment from 92.83\% to 39.34\%, demonstrating the importance
of independently timed text conditioning for shot and dialogue
scheduling. The directorial lip-sync mechanism addresses a
complementary aspect of control: whether active speech should
drive visible articulation. Compared with native audiovisual
coupling, it reduces listener TemporalBias from 1.78 to 0.33,
an 81.5\% relative reduction, while Imaging Quality, speaker Sync-C, and audio metrics remain broadly stable. These results indicate reduced unwanted
audiovisual coupling on listener shots while retaining
visible-speaker synchronization. A matched qualitative
comparison is provided in Appendix~\ref{app:controller-case},
with videos in the supplementary materials. To assess the computational cost of these gains, we compare
inference runtime with and without the directorial lip-sync
mechanism using the same model checkpoint. On a single
NVIDIA H20 GPU, enabling the mechanism increases the per-step
runtime from 12.81 to 13.01 seconds
(Table~\ref{tab:ablation}), corresponding to an additional
0.20 seconds (1.56\%). This modest overhead requires no
additional training or learned parameters.

\begin{table}[t]
\vspace{-10pt}
  \centering
  \footnotesize
  \setlength{\tabcolsep}{0pt}
  \renewcommand{\arraystretch}{1.15}
  \begin{tabular*}{\linewidth}{@{\extracolsep{\fill}} l cc cccc cc c @{}}
    \toprule
    & \multicolumn{2}{c}{Video Quality}
    & \multicolumn{4}{c}{Audio Quality}
    & \multicolumn{2}{c}{AV Alignment}
    & Efficiency \\
    \cmidrule(lr){2-3} \cmidrule(lr){4-7} \cmidrule(lr){8-9} \cmidrule(lr){10-10}
    Configuration
    & Cut F1 $\uparrow$ & Imaging $\uparrow$
    & WER $\downarrow$ & Cont. $\uparrow$ & Leak $\downarrow$ & Sim. $\uparrow$
    & Bias (L) $\downarrow$ & Sync-C (S) $\uparrow$ & Time (s/step) $\downarrow$ \\
    \midrule
    \multicolumn{10}{@{}l}{\textit{Dual-schedule temporal CA}} \\
    w/o Temporal CA & 0.34 & 0.65 & 6.57 & 39.34 & 25.98 & 0.39 & 2.31 & 3.16 & \textbf{12.81} \\
    \midrule
    \multicolumn{10}{@{}l}{\textit{Directorial Lip-Sync Mech.}} \\
    w/o Mechanism & \textbf{0.99} & \textbf{0.66} & 4.38 & 90.89 & \textbf{0.74} & 0.55 & 1.78 & \textbf{3.53} & \textbf{12.81} \\
    w/ Mechanism (\textbf{Ours}) & \textbf{0.99} & \textbf{0.66} & \textbf{4.16} & \textbf{92.83} & 0.77 & \textbf{0.56} & \textbf{0.33} & 3.42 & 13.01 \\
    \bottomrule
  \end{tabular*}
  \caption{Ablation study on core components. Evaluated on a fixed subset of the 15-second conversational benchmark. Best values within each block are in \textbf{bold}. Notably, the lip-sync mechanism is entirely training-free: both configurations in the second block operate on the exact same checkpoint. Runtime denotes seconds per denoising step measured on a single NVIDIA H20 GPU.}
  \label{tab:ablation}
\vspace{-10pt}
\end{table}

\section{Conclusion}
\label{sec:conclusion}

\decoupage{} enables independent handling of visual shot cuts and continuous cinematic audio through dual‑schedule temporal cross‑attention. It orchestrates shot transitions and dialogue schedules via decoupled time schedules, while the directorial lip‑sync mechanism leverages auxiliary non‑speech acoustic states to suppress phantom articulation on silent listeners. Together, these designs naturally unlock cinematic reaction shots and sound‑bridge effects for long‑form video generation. Evaluations over 15‑second and 60‑second multi‑shot conversational scenes demonstrate clear improvements in cut adherence, dialogue containment, and recurring‑character voice consistency, yielding high‑quality narrative multi‑shot audiovisual outputs.
\paragraph{Limitations and Future Work.}
Our current formulation targets turn‑taking dialogue and does not explicitly support simultaneous, overlapping speech from multiple characters. This limits performance for crowded group scenes where several actors speak concurrently. One potential path forward is to extend our acoustic track to model multiple co‑active speech events within the same time interval. Moving forward, an exciting frontier is exploring real‑time, directorial‑level joint audiovisual generation to enable responsive, on‑the‑fly cinematic creation. In parallel, we envision developing assisted screenplay authoring tools to simplify high‑level creative control and lower end‑user barriers.
\clearpage
\section*{AI use statement}

Generative AI tools were used to assist with literature organization and the drafting and revision of manuscript text, including improving clarity, grammar, and readability. All AI-assisted content was reviewed and edited by the authors to ensure that it accurately reflected the research conducted. The authors verified the cited sources, checked technical descriptions and implementation claims against the code, and reviewed the experiments and results included in the paper. All substantive research decisions and interpretations of the findings remained the responsibility of the authors. The authors take full responsibility for the accuracy, originality, and integrity of the final manuscript.

\section*{Reproducibility Statement}
To support reproducibility, we will release the full codebase and accompanying documentation upon publication. The release will include environment specifications and dependencies, model configurations, hyperparameters, random seeds, data preparation and splitting procedures, and training schedules. We will also provide the evaluation scripts and metric implementations used to produce the reported results, together with step-by-step commands for running the experiments. Hardware configurations and execution-time information will be documented to help researchers assess the computational requirements. These materials will enable independent verification of our findings and facilitate further research building on our work.

\bibliography{paper}
\bibliographystyle{iclr2027_conference}

\appendix
\clearpage
\section{Supplementary Results}
\subsection{Minute-Long Qualitative Results}
\label{app:60s}
\begin{figure}[t]
\vspace{-16pt}
    \centering
    \includegraphics[width=\textwidth]{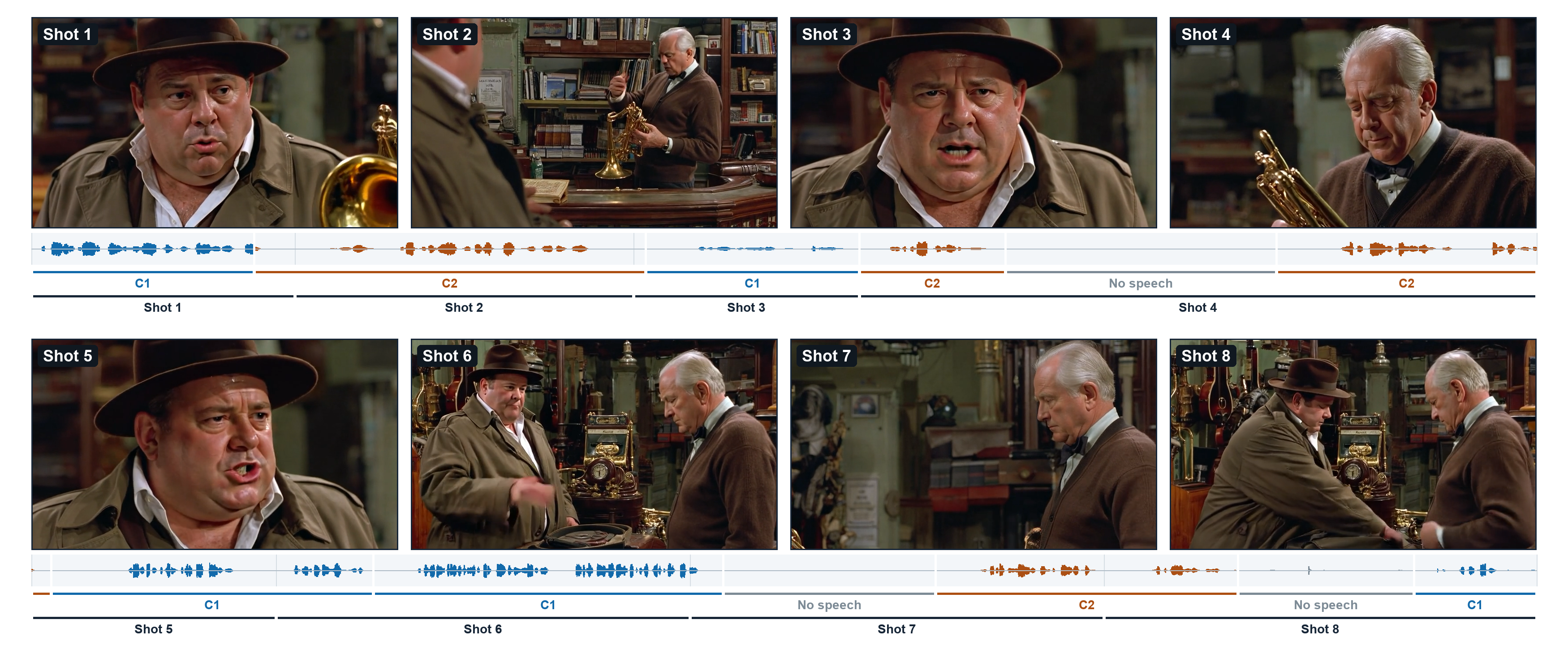}
    \caption{Qualitative example of a 60-second generation. Eight representative frames show a two-character exchange unfolding across multiple shots in a trumpet shop. Beneath each row, the generated audio waveform is aligned with the corresponding shot sequence. The colored strips indicate the screenplay-specified speaking role (C1 or C2), while gray segments denote intervals with no requested speech. The example visualizes coordinated shot transitions, character turns, and speech--silence timing over a full-minute sequence.}
    \label{fig:case0039-storyboard-audio}
\end{figure}

Figure~\ref{fig:case0039-storyboard-audio} illustrates our model's ability to generate a 60-second multi-shot audiovisual sequence. More examples are provided in the supplementary materials.

\subsection{Metric Definitions and Evaluation Protocol}
\label{app:evaluation-details}

\noindent\textbf{Dialogue accuracy.}
For video $i$, let $n_i$ be the number of reference words and let $s_i$, $d_i$,
and $u_i$ denote substitutions, deletions, and insertions between the screenplay
dialogue and the Faster-Whisper transcript. We report corpus-level word error
rate as
\begin{equation}
\operatorname{WER}
=100\,\frac{\sum_i(s_i+d_i+u_i)}{\sum_i n_i}.
\label{eq:corpus-wer}
\end{equation}

\noindent\textbf{Silence leakage.}
Let $S_i$ be the union of requested silence intervals in video $i$,
$S_i^{-\tau}$ the union after removing $\tau$ seconds from each end of each
interval, and $W_i$ the union of recognized word intervals. With $\mu$
denoting duration, we define
\begin{equation}
\operatorname{SL}_{\tau}
=100\,
\frac{\sum_i\mu\!\left(S_i^{-\tau}\cap W_i\right)}
     {\sum_i\mu\!\left(S_i\right)},
\qquad \tau=0.2\,\mathrm{s}.
\label{eq:silence-leak}
\end{equation}
Intervals shorter than $2\tau$ have no remaining interior, while the
denominator retains the original requested silence duration. Using word
intervals avoids counting pauses enclosed within longer recognition segments.

\noindent\textbf{Text-aligned containment.}
For turn $t$, let $[a_t,b_t]$ be the allowed speech window and
$[\hat a_t,\hat b_t]$ its boundaries obtained from transcript alignment. The
total boundary overflow is
\begin{equation}
o_t=\max(0,a_t-\hat a_t)+\max(0,\hat b_t-b_t).
\label{eq:turn-overflow}
\end{equation}
For the set $T$ of all target turns and its validly aligned subset $A$,
containment at tolerance $\delta$ is
\begin{equation}
\operatorname{Containment}_{\delta}
=\frac{100}{|T|}\sum_{t\in A}\mathbf{1}[o_t\leq\delta].
\label{eq:text-containment}
\end{equation}
Missing or invalid alignments remain in the denominator and count as failures;
unused time within the allowed window incurs no penalty. We use
$\delta=0.2\,\mathrm{s}$ for the 15-second evaluation dataset and
$\delta=0.4\,\mathrm{s}$ for the 60-second evaluation dataset.
\begin{figure}[t]
\vspace{-5pt}
    \centering
    \includegraphics[width=\textwidth]{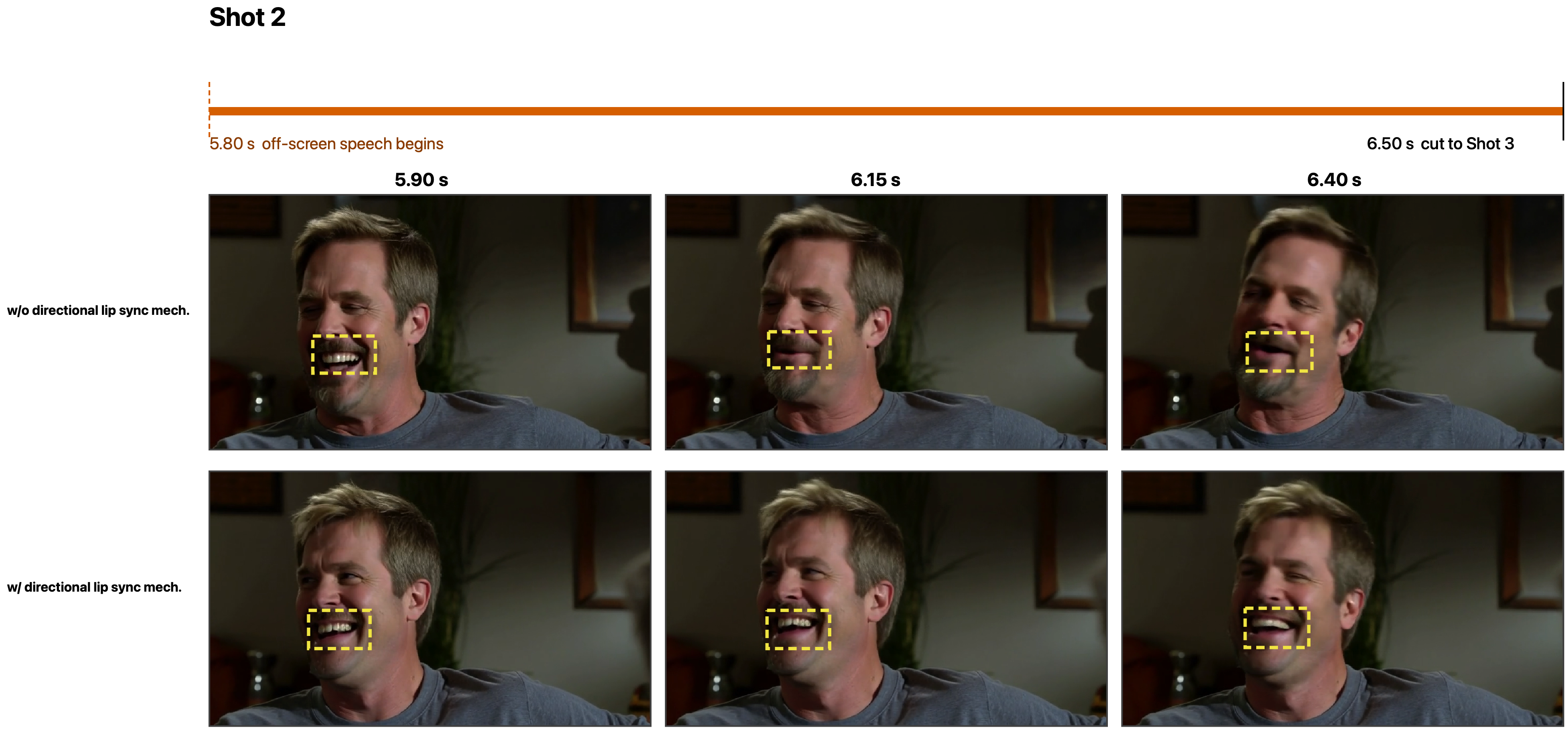}
    \caption{Matched comparison of the Directorial Lip-Sync Mechanism near the end of Shot~2. The first row shows results without the mechanism, and the second row shows results with it, at 5.90, 6.15, and 6.40 seconds. Yellow dashed boxes highlight the mouth region.}
    \label{fig:dbecd-shot2-tail-controller}
\vspace{-7.5pt}
\end{figure}
\noindent\textbf{Recurring-role voice consistency.}
Let $e_{ict}$ be the frozen ECAPA-TDNN embedding of an eligible turn $t$ for
character $c$ in video $i$. Let $P_{ic}$ contain all unordered pairs of
distinct eligible turns for that character, $C_i$ the characters with at least
one such pair, and $\mathcal V$ the videos with at least one eligible
character. We compute
\begin{equation}
\operatorname{Consistency}
=\frac{1}{|\mathcal V|}\sum_{i\in\mathcal V}
 \frac{1}{|C_i|}\sum_{c\in C_i}
 \frac{1}{|P_{ic}|}\sum_{(t,u)\in P_{ic}}
 \frac{e_{ict}^{\top}e_{icu}}
      {\lVert e_{ict}\rVert_2\lVert e_{icu}\rVert_2}.
\label{eq:voice-consistency}
\end{equation}
Timestamp-window crops use the requested speech windows and their screenplay
character labels. Role Margin is computed within each eligible video as the
difference between same-role and different-role similarity, rather than by
subtracting independently supported aggregate means.

\noindent\textbf{Audio-Video Alignment}
For Audio-Video alignment, prompt speech timestamps are
intersected with requested shot intervals and visible roles to form listener
and visible-speaker windows of at least 2.2 seconds. A single S3FD track must
cover at least 95\% of each continuous window. Imaging Quality is computed on complete 15-second clips. Shot
transitions are detected at a fixed threshold of 0.18 and matched one-to-one to
requested cuts within 0.5 seconds.

\paragraph{TemporalBias formulation.}
To measure soundtrack-specific temporal coupling on listener/off-screen windows,
we define an Audio-Lock contrast using SyncNet confidence. For each window $m$,
let $C_m(v_m,a_m)$ denote the near-lag SyncNet confidence between the detected
facial track $v_m$ and its aligned audio $a_m$. We construct temporally displaced
controls by circularly shifting $a_m$ by a fraction $\delta$ of the window
duration, where $\mathcal D=\{0.25,0.375,0.5,0.625\}$. The contrast is
\begin{equation}
\Delta C_m
=
C_m(v_m,a_m)
-
\underset{\delta\in\mathcal D}{\operatorname{median}}
C_m\!\left(v_m,\operatorname{shift}(a_m,\delta)\right).
\end{equation}
We then aggregate the contrast magnitude over the $N$ listener/off-screen
windows:
\begin{equation}
\operatorname{TemporalBias}
=
\frac{1}{N}\sum_{m=1}^{N}|\Delta C_m|.
\end{equation}
If a listener's facial dynamics exhibit no soundtrack-specific temporal
preference, $\Delta C_m$ should be close to zero. Accordingly, a lower
TemporalBias indicates weaker unwanted audiovisual temporal coupling and is
better.

\subsection{Matched Qualitative Directional Lip-Sync Mechanism Case}
\label{app:controller-case}

Figure~\ref{fig:dbecd-shot2-tail-controller} compares matched trajectories without and with the directional Lip-sync mechanism at three aligned timestamps near the end of
Shot~2. The prompt, seed, checkpoint, sampling settings, and native V2A path
are fixed; only the inference-time mechanism setting differs.
Character~2 remains visible while Character~1 speaks off screen from 5.80
seconds until the cut at 6.50 seconds. All three displayed timestamps lie
inside this interval. In these frames, the output without the mechanism moves
from an open laugh toward a narrower mouth configuration during the off-screen
line, whereas the matched output with the mechanism remains in an open-laugh
configuration. Videos can be found in the supplementary materials.

\end{document}